\documentclass[lettersize,journal]{IEEEtran}
\usepackage{hyperref}

\usepackage{amsmath,amsfonts}
\usepackage{algorithmic}
\usepackage{algorithm}
\usepackage{array}
\usepackage{textcomp}
\usepackage{stfloats}
\usepackage{url}
\usepackage{verbatim}
\usepackage{graphicx}
\usepackage{kotex}
\usepackage{amsmath,amsfonts}
\usepackage{algorithmic,algorithm}
\usepackage{array}
\usepackage[caption=false,font=normalsize,labelfont=sf,textfont=sf]{subfig}
\usepackage{textcomp,stfloats,url,verbatim,graphicx,kotex}

\usepackage[table,dvipsnames]{xcolor}      % \cellcolor, \rowcolor
\usepackage{booktabs,multirow,colortbl,adjustbox}
\usepackage{pifont}                        % \ding 기호

\usepackage{etoolbox}   % for \begingroup … \endgroup trick
\usepackage{stfloats}           % 두-컬럼 bottom float 등을 허용
\usepackage{placeins}   % \FloatBarrier 를 제공
\usepackage{array, colortbl, xcolor, multirow,tabularx}


\usepackage[ % IEEE style citation for biblatex
style=ieee, backend=biber, sorting=none,
maxcitenames=3, mincitenames=1, maxnames=3, minnames=1,
]{biblatex} % Use the biblatex package for citation, replacing the `cite` package.
\DefineBibliographyStrings{english}{
  in = {in \it{Proc}\adddot},
}
\newcommand{\degc}[1]{$^\circ$C} % Textcolor blue macro

\usepackage{kotex} % For Korean language
\usepackage{soul} % For strikeout the text by \st{}
\usepackage{svg}

\begin{document}

\title{UM-Depth : Uncertainty Masked Self-Supervised Monocular Depth Estimation with Visual Odometry}

\author{
  Tae-Wook Um, Ki-Hyeon Kim, Hyun-Duck Choi and Hyo-Sung Ahn
  % <-this % stops a space
  \thanks{This work was supported by Korea Institute of Planning and Evaluation for Technology in Food, Agriculture and Forestry(IPET) through High Value-added Food Technology Development Program, funded by Ministry of Agriculture, Food and Rural Affairs(MAFRA)(RS-2022-IP322054)}% <-this % stops a space
  \thanks{This work was supported by the National Research Foundation of Korea (NRF) grant funded by the Korea government (MSIT) (2022R1A2B5B03001459).}
  \thanks{Tae-Wook Um, Ki-Hyeon Kim and Hyo-Sung Ahn are with the School of Mechanical and Robotics Engineering, Gwangju Institute of Science and Technology (GIST), Gwangju 61005, South Korea (e-mail: taewookum@gm.gist.ac.kr; rlgus1394@gm.gist.ac.kr; hyosung@gist.ac.kr).}
  \thanks{Hyun-Duck Choi is with Department of Smart ICT Convergence Engineering, Seoul National University of Science and Technology, Seoul 01811, South Korea (e-mail: ducky.choi@seoultech.ac.kr).}
  \thanks{(Corresponding author: Hyo-Sung Ahn.)}
}

\markboth{Journal of \LaTeX\ Class Files,~Vol.~14, No.~8, August~2021}%
{Shell \MakeLowercase{\textit{et al.}}: A Sample Article Using IEEEtran.cls for IEEE Journals}%Uncertainty Mask Self Supervised Depth Estimation

% \IEEEpubid{0000--0000/00\$00.00~\copyright~2021 IEEE}-이거 일단 주석 처리

\maketitle

\begin{abstract}
    Monocular depth estimation has been increasingly adopted in robotics and autonomous driving for its ability to infer scene geometry from a single camera. In self-supervised monocular depth estimation frameworks, the network jointly generates and exploits depth and pose estimates during training, thereby eliminating the need for depth labels.
    However, these methods remain challenged by uncertainty in the input data, such as low-texture or dynamic regions, which can cause reduced depth accuracy.
    To address this, we introduce UM-Depth, a framework that combines motion- and uncertainty-aware refinement to enhance depth accuracy at dynamic object boundaries and in textureless regions.
    Specifically, we develop a teacher-student training strategy that embeds uncertainty estimation into both the training pipeline and network architecture, thereby strengthening supervision where photometric signals are weak.
    Unlike prior motion-aware approaches that incur inference-time overhead and rely on additional labels or auxiliary networks for real-time generation, our method uses optical flow exclusively within the teacher network during training, which eliminating extra labeling demands and any runtime cost.
    Extensive experiments on the KITTI and Cityscapes datasets demonstrate the effectiveness of our uncertainty-aware refinement. Overall, UM-Depth achieves state-of-the-art results in both self-supervised depth and pose estimation on the KITTI datasets.
\end{abstract}

\begin{IEEEkeywords}
  Monocular depth estimation, Uncertainty analysis,
  State space methods, Optical flow, Visual odometry
\end{IEEEkeywords}

\section{Introduction}
\IEEEPARstart{M}{onocular} depth estimation (MDE), defined as the recovery of 3-D geometry from a single 2-D image, underpins a wide range of computer-vision applications, including autonomous driving~\cite{geiger2012we}, augmented reality~\cite{swan2006perceptual}, and 3-D rendering~\cite{bonatto2021real}. Because it requires only a single camera, MDE provides a cost-effective solution with reduced hardware complexity and easier system integration compared to stereo or LiDAR-based systems.
Early learning-based MDE methods relied on fully supervised training with depth ground truth captured by active sensors (e.g., LiDAR or time-of-flight cameras)~\cite{eigen2014depth}.

Because collecting such data is costly and logistically demanding, recent research has shifted toward self-supervised learning, in which the network is trained by minimizing the photometric reconstruction error between adjacent frames~\cite{godard2017unsupervised,zhan2018unsupervised}.
Self-supervised frameworks can be broadly divided into stereo-based and monocular-video-based paradigms.
Stereo-based approaches~\cite{garg2016unsupervised,godard2017unsupervised} dispense with an explicit pose-estimation network, but they are highly sensitive to precise baseline distances and accurate camera intrinsics; even small calibration errors can severely degrade depth predictions, making data preparation almost as cumbersome as in supervised settings.
Monocular-video-based approaches~\cite{zhou2017unsupervised,zhan2018unsupervised,} typically incorporate an additional PoseNet to estimate the relative camera motion between successive frames, which simplifies data acquisition because only a single moving camera is needed.
Nevertheless, in both stereo- and monocular-video frameworks, the network purely relies on photometric supervision with a static scene assumption. This assumption is violated by moving objects, which cause inconsistent pixel correspondences, while texture-deficient regions yield ambiguous photometric signals. Both factors contribute to increased prediction uncertainty.
To mitigate these limitations, recent studies aggregate temporal geometry across multiple frames, typically by building cost volumes or performing feature matching, to improve consistency in static regions~\cite{watson2021temporal,feng2022disentangling}.
A complementary line of work explicitly estimates per-pixel uncertainty~\cite{poggi2020uncertainty} and down-weights the photometric loss in high-uncertainty areas, thereby reducing the impact of noisy observations. Although these strategies improve robustness, they still fall short of fully addressing the uncertainty introduced by moving objects and severely texture-less surfaces in real-world scenes.

Previous attempts to handle non-static scenes have incorporated edge-aware smoothness terms to enforce depth continuity near object boundaries~\cite{godard2017unsupervised} and segmentation-based masking of dynamic regions to refine depth around moving objects~\cite{chen2023self,jung2021fine}.
However, edge-aware methods do not explicitly distinguish moving objects, while segmentation-based approaches require additional labeled data and preprocessing.
We address these weaknesses with \textbf{UM-Depth}, a self-supervised framework that augments existing motion-aware and multi-frame strategies with an explicit, uncertainty-guided refinement mechanism organized in a teacher-student configuration.
Our teacher network employs a single-frame DepthNet, PoseNet, and FlowNet to generate depth, camera pose, and optical flow estimates. These estimates serve as guidance signals, enabling us to identify dynamic regions with a motion-aware manner, without any kind of additional metadata (e.g., optical flow) or inference-time cost.
The student network applies a isolated triplet loss~\cite{chen2023self} to intermediate depth features, guided by the teacher’s motion-aware signals, to enhance depth estimation around dynamic objects.
The student network was built upon the multi-frame design of ManyDepth~\cite{watson2021temporal}, which integrates a Mamba-based encoder~\cite{gu2023mamba,shaker2024groupmamba} and a PBU-HRNet decoder equipped with learnable prompts and adaptive depth bins. These architectural components efficiently capture both global and local information, adapting to scene-specific geometric distributions.

Experimental results confirm that the proposed method overcomes the main weaknesses of prior self-supervised MDE approaches and attains state-of-the-art accuracy on the KITTI Depth \cite{geiger2012we} and KITTI Odometry \cite{geiger2013vision} benchmarks, while also delivering competitive performance on Cityscapes \cite{cordts2016cityscapes}.

\subsection*{Contributions}
This work advances self-supervised monocular depth estimation in four ways:

\begin{itemize}
      \item We introduce a teacher–student framework that detects dynamic regions in a motion-aware manner and guides the student with a isolated triplet loss, improving depth around moving-object boundaries without extra inference cost.

      \item The student employs a Mamba encoder to aggregate temporal context and a PBU-HRNet decoder to capture spatial detail, improving both accuracy and efficiency over prior CNN and Transformer baselines.

      \item The model predicts per-pixel uncertainty maps to identify low-confidence regions. These regions are refined using a lightweight range-map module, leading to better predictions in texture-less or dynamic areas.

      \item UM-Depth achieves state-of-the-art depth and pose accuracy on the KITTI Depth and KITTI Odometry benchmarks, while maintaining competitive performance on Cityscapes. Ablation studies confirm the effectiveness of each proposed component.
\end{itemize}

\section{Related Work}

\subsection{Self-Supervised Monocular Depth Estimation}
Estimating depth from a single image is inherently challenging, because one 2-D view can correspond to multiple plausible 3-D scenes. To address this ambiguity, numerous deep-learning approaches have been proposed. Early work by Eigen \emph{et al.}~\cite{eigen2014depth} adopted fully supervised learning and directly regressed per-pixel depth values; although accurate, the method depends on dense ground-truth labels, which are costly and labor-intensive to acquire. To remove this constraint, Garg \emph{et al.}~\cite{garg2016unsupervised} introduced an unsupervised scheme that employs view synthesis on stereo pairs, yet the approach assumes known camera poses and static scenes, restricting real-world applicability. Subsequently, Zhou \emph{et al.}~\cite{zhou2017unsupervised} presented a framework that jointly trains a depth network and a pose network on monocular video; their system conducts view synthesis between consecutive frames and optimizes the networks by minimizing photometric reconstruction error. Building on this idea, Godard \emph{et al.}~\cite{godard2019digging} proposed an auto-masking strategy that ignores pixels with unreliable photometric errors, thereby mitigating occlusion effects.
Meanwhile, several studies have leveraged multi-view input, typically adjacent video frames from a monocular camera, to further enhance depth estimation. These methods construct cost volumes or perform feature matching across views to exploit parallax, and detect dynamic objects by analyzing motion inconsistencies over time. Such strategies have yielded significant gains in monocular depth accuracy~\cite{feng2022disentangling,watson2021temporal}. In this study, we extend ManyDepth~\cite{watson2021temporal} by inserting a memory-efficient Mamba encoder, which processes long frame sequences with linear complexity and improves both runtime efficiency and depth accuracy.

\subsection{Uncertainty}
Depth estimation performan ce often degrades in regions where prediction uncertainty is high, such as low-texture areas or dynamic objects. This uncertainty arises when the model cannot confidently resolve depth due to insufficient visual cues (e.g., low-texture regions) or conflicting information (e.g., dynamic objects). In such cases, predictions tend to be noisy or biased, leading to spatial inconsistencies and degraded geometric accuracy.
It is typically categorized into \textit{epistemic uncertainty}, which arises from model parameters, and \textit{aleatoric uncertainty}, which originates from intrinsic data noise or ambiguity. Kendall \emph{et al.}~\cite{kendall2017uncertainties} employed a Bayesian neural network to capture epistemic uncertainty while simultaneously learning aleatoric uncertainty, which is often high in low-texture regions such as sky or walls. Their model estimates per-pixel variance and assigns lower weights to high-uncertainty pixels in the loss function, thereby improving depth quality.

Within self-supervised settings, where ground-truth depth is unavailable, Li \emph{et al.}~\cite{li2023uncertainty} propose generating multi scale depth maps and quantifying uncertainty by measuring their disparities; this uncertainty is then integrated into the loss as a weighting factor. Likewise, Poggi \emph{et al.}~\cite{poggi2020uncertainty} introduce a teacher–student framework that compares the two networks’ outputs to estimate uncertainty, enabling effective use of confidence maps without any labeled data.

\subsection{Prompt-Based Depth Estimation}
Prompting strategies have recently received attention as a means of adapting vision models to specific tasks. Early studies incorporated external text prompts, distinct from the input images, to steer model training~\cite{zhou2022learning,radford2021learning}; however, these methods inherently depend on a text encoder. To remove this dependency, Jia \emph{et al.}~\cite{jia2022visual} devised a technique that applies prompts directly to the input image, thereby eliminating textual overhead.

Building on this foundation, subsequent works~\cite{potlapalli2023promptir,zhu2023visual,wang2024tsp,jia2022visual} explore \emph{learnable} visual prompts that guide models toward task-relevant features during training, allowing efficient specialization without extensive architectural modification. Motivated by these advances, we adopt a novel prompting mechanism that enriches image representations required for depth estimation while maintaining network simplicity.

\subsection{Mamba Architecture}

Mamba, a recent state space model architecture, has emerged as a promising alternative to Transformer- and CNN-based architectures due to its ability to capture long-range dependencies within sequences~\cite{gu2023mamba}. For instance, Liu \emph{et al.}~\cite{liu2024vmamba} leverage Mamba blocks and demonstrate excellent performance on long-sequence modeling benchmarks.
In vision, interest in Mamba is rapidly growing. For classification, approaches such as~\cite{liu2024vmamba,shaker2024groupmamba} introduce multi-directional scanning to globally extend receptive fields. In addition, segmentation frameworks employing Mamba-based modules attain high accuracy~\cite{liu2024swin}. More recently, hybrid architectures that combine CNNs, Vision Transformers (ViTs), and Mamba, as exemplified by~\cite{hatamizadeh2024mambavision}, have proven effective at simultaneously capturing local and global dependencies. In the context of depth estimation, long-range context is especially useful for resolving ambiguous regions such as low-texture surfaces or occluded objects, where local information alone may be insufficient to infer accurate depth. Temporal cues across multiple frames often span large spatial distances, further motivating the need for models with extended receptive fields.
Inspired by these findings, our method employs a Mamba encoder to model long-range context while preserving high-resolution details through a complementary decoding stage.

\section{Method}
In this section, we introduce the framework of self-supervised monocular depth estimation. Our proposed model consists of a single-frame teacher model, a multi-frame student model, and an optical flow model. We describe the student network's encoder, which is based on Mamba~\cite{gu2023mamba,shaker2024groupmamba}, and its decoder, which utilizes prompts. Finally, we introduce a dynamic object masking technique using optical flow and a depth refinement approach based on uncertainty estimation.

\subsection{Self-Supervised Monocular Depth Estimation}
\begin{figure*}[t]
  \centering
  \includegraphics[width=1.0\textwidth]{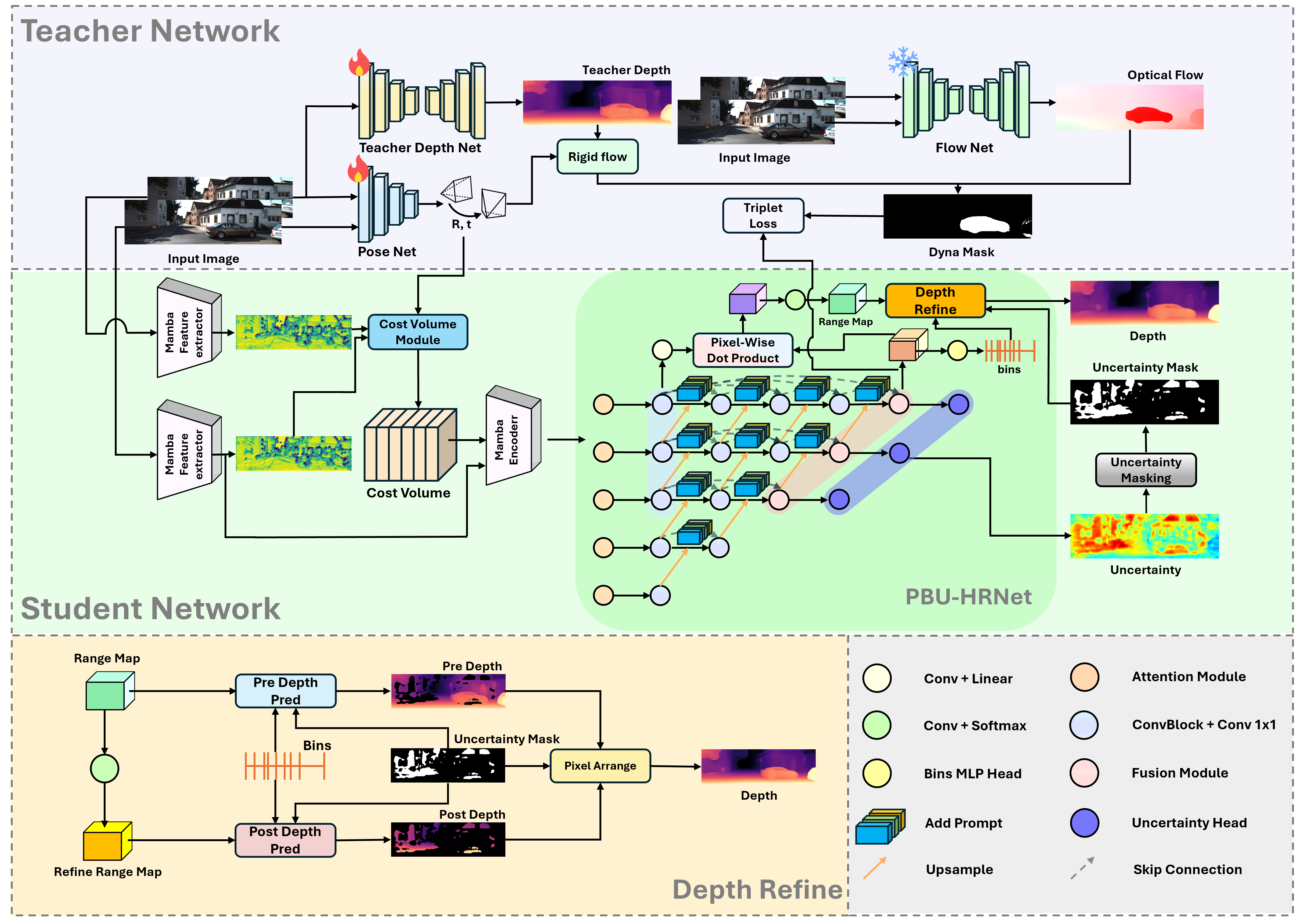}
  \vspace{-0.6cm}
  \caption{An illustration of the proposed UM-Depth network.
    (a) The teacher network consists of a convolution-based single-frame depth network and an optical flow network, providing supervision signals for depth and motion estimation. (b) The student network is a multi-frame depth estimator that builds a cost-volume, processes it with a GroupMamba encoder, and decodes the features with our Prompt, Bins, and Uncertainty HRNet (PBU-HRNet). PBU-HRNet combines a learnable-prompt branch (for richer semantics) with a bins-based uncertainty branch that discretizes depth, estimates per-bin probabilities, and refines pixels with high uncertainty. The resulting uncertainty mask selectively replaces unreliable predictions, boosting overall depth accuracy.
  }
  \label{fig:um_model}
\end{figure*}

Following the prior work~\cite{watson2021temporal}, the goal is to estimate per-pixel depth without ground truth by utilizing a depth network and a pose network given a target image $I_{t}$ and source images $I_{s}$, where $s \in \{t-1, t+1\}$. Our approach is designed based on the architecture of ManyDepth, where the student network estimates the depth map $D_{t}$ using multi-frames:
\begin{equation}
  D_{t} = \theta_{\text{multi}}(I_{t}, I_{t-1}),
  \label{eq:depth_estimation}
\end{equation}
where $\theta_{\text{multi}}$ represents the multi-frame depth estimation network. Moreover, we employ a pose network based on ResNet~\cite{he2016deep}, which takes the target image $I_{t}$ and source image $I_{s}$ as inputs to estimate the relative pose $T_{t \to s}$:
\begin{equation}
  T_{t \to s} = \theta_{\text{pose}}(I_{t}, I_{s}),
  \label{eq:pose_estimation}
\end{equation}
where $\theta_{\text{pose}}$ denotes the pose estimation network.
Using the depth map of the target image $D_{t}$ and the relative pose $T_{t \to s}$, the source image $I_{s}$ is mapped to the target image $I_{t}$ to obtain the synthesized image $I_{s \to t}$~\cite{zhou2017unsupervised}:
\begin{equation}
  I_{s \to t} = I_{s} \langle K T_{t \to s} D_{t} K^{-1} p_{t} \rangle,
  \label{eq:image_synthesis}
\end{equation}
where $p_{t}$ denotes the homogeneous coordinates of $I_{t}$, and $K$ represents the camera intrinsic matrix, which is assumed to be known. The operator $\langle \cdot \rangle$ denotes the sampling operator.
To compute the reconstruction loss between the synthesized image and the target image, we employ the photometric error $pe(\cdot)$~\cite{godard2017unsupervised}, which combines the $L_{1}$ loss and the structural similarity index measurement (SSIM) loss ~\cite{wang2004image}:
\begin{equation}
  pe(I_{s \to t}, I_{t}) = \frac{\alpha}{2} (1 - \text{SSIM}(I_{s \to t}, I_{t})) + (1 - \alpha) \lVert I_{s \to t} - I_{t} \rVert,
  \label{eq:photometric_error}
\end{equation}
where $\alpha$ is set to 0.85. To optimize the photometric loss~\cite{godard2019digging} for each pixel, we take the minimum across source views:
\begin{equation}
  L_{\text{ph}} = \min_{s} pe(I_{s \to t}, I_{t}).
  \label{eq:photometric_loss}
\end{equation}
To ensure the smoothness of the estimated depth map, we employ the edge-aware smoothness loss proposed in~\cite{godard2017unsupervised}:
\begin{equation}
  L_{\text{sm}} = \lvert \partial_{x} d^{*}_{t} \rvert e^{-\lvert \partial_{x} I_{t} \rvert} + \lvert \partial_{y} d^{*}_{t} \rvert e^{-\lvert \partial_{y} I_{t} \rvert},
  \label{eq:smoothness_loss}
\end{equation}
where $d_{t}^{*} = d_{t} / \bar{d}_t$ normalizes inverse depth by its image-wise mean $\bar{d}_t$, eliminating scale ambiguity so the smoothness loss penalizes only relative gradients.
Following ManyDepth~\cite{watson2021temporal}, the teacher’s depth network $\theta_{\text{single}}$ and the student’s multi-frame cost-volume network share the same camera pose, ensuring scale-consistent predictions (see Fig.~\ref{fig:um_model} for the multi-frame cost-volume architecture).
To stabilize training, we isolate unreliable pixels using a binary inconsistency mask $M$, because the student’s cost volume often fails to generalize in dynamic or low-texture regions. To construct this mask, we compare the teacher’s predicted depth $\hat{D}_{t}$ with $D_{\text{cv}}$, a depth map derived from the student’s cost volume.
The cost volume encodes matching costs across discretized depth hypotheses, and $D_{\text{cv}}$ is obtained by selecting, at each pixel, the depth corresponding to the minimum matching cost (i.e., via an $\arg\min$ operation over the depth dimension). A pixel is masked as inconsistent ($M{=}1$) if the maximum of the two relative differences, between $\hat{D}_{t}$ and $D_{\text{cv}}$, exceeds a threshold of 1, as defined in~\eqref{eq:mask}:

\begin{equation}
  M = \max \left( \frac{D_{\text{cv}} - \hat{D}_{t}}{\hat{D}_{t}}, \frac{\hat{D}_{t} - D_{\text{cv}}}{D_{\text{cv}}} \right) > 1.
  \label{eq:mask}
\end{equation}

For the masked regions, an $L_{1}$ loss is applied to $D_{t}$ to encourage consistency with $\hat{D}_{t}$, as expressed in~\eqref{eq:consistency_loss}:

\begin{equation}
  L_{\text{consistency}} = \sum M \lvert D_{t} - \hat{D}_{t} \rvert.
  \label{eq:consistency_loss}
\end{equation}

Pixels outside the mask are supervised using the standard photometric loss $L_{\text{ph}}$, while masked pixels receive additional guidance through the consistency loss $L_{\text{consistency}}$ as defined in~\eqref{eq:consistency_loss}. During backpropagation, gradients are blocked through the teacher’s output $\hat{D}_t$ to ensure that the consistency loss influences only the student network. While the teacher can still be trained independently (e.g., via its own photometric loss), it remains unaffected by the consistency supervision. This selective gradient blocking preserves the teacher’s role as a stable guidance signal during training. The overall training objective incorporating these terms is defined in~\eqref{eq:baseline_loss}:

\begin{equation}
  L_{\text{self}} = (1 - M) L_{\text{ph}} + L_{\text{consistency}} + \lambda_{\text{sm}} L_{\text{sm}},
  \label{eq:baseline_loss}
\end{equation}
where $\lambda_{\text{sm}}$ is the weight for the smoothness loss, set to $10^{-3}$.

\subsection{Dynamic Object Masking}
Depth estimation accuracy tends to degrade near object boundaries, particularly for dynamic objects, which are a major source of performance deterioration in self-supervised depth estimation. To mitigate this issue, prior works have introduced edge-aware smoothness loss~\cite{godard2017unsupervised} to enforce the continuity of depth maps. An alternative approach is to leverage segmentation images, as proposed by~\cite{chen2023self, jung2021fine}, to explicitly delineate object boundaries. These methods use segmentation masks to guide auxiliary losses such as triplet loss  $L_{\text{tri}}$, encouraging the network to sharpen depth discontinuities at object edges. While effective, segmentation-based approaches~\cite{chen2023self, jung2021fine} require either precomputed labels or the design of an additional network, increasing training complexity and data preparation costs, thereby limiting their practicality in self-supervised settings.

To address this limitation, we propose a method that identifies dynamic objects via optical flow. Unlike prior approaches~\cite{sun2024flowdepth}, which utilizes optical flow during both training and inference, our method integrates the optical flow module into the teacher network and uses it exclusively during training.
This allows the depth network to benefit from motion-aware supervision without introducing any optical flow-related cost at inference time. The generated binary mask highlights dynamic regions, where we apply a triplet loss that helps the network better distinguish depth differences between foreground and background by grouping pixels with similar motion and separating those with different motion patterns. Compared to segmentation-based approaches, which rely on predefined object classes and labeled data, our method leverages motion information to detect a wider range of dynamic regions without requiring manual annotations.

We employ RPKNet~\cite{morimitsu2024recurrent} in the teacher network to estimate optical flow between consecutive frames. The optical flow from frame $t$ to $t{-}1$ is computed as:
\begin{equation}
  F_{t \to t-1} = \theta_{\text{flow}}(I_{t}, I_{t-1}),
  \label{eq:flow_estimation}
\end{equation}
where $\theta_{\text{flow}}$ denotes the optical flow network and $I_{t}$ and $I_{t-1}$ are the input images.

In parallel, we compute a rigid flow $\hat{F}_{t \to t-1}$ based on the teacher's depth prediction and relative camera pose. This flow assumes that the scene is entirely static and that all pixel motion is caused solely by camera egomotion. The rigid flow is given by:
\begin{equation}
  \hat{F}_{t \to t-1} = K \bigl( T_{t \to t-1} \hat{D}_{t} K^{-1} p_{t} \bigr) - p_{t},
  \label{eq:2d_motion}
\end{equation}
where $K$ is the camera intrinsics, $\hat{D}_{t}$ is the predicted depth map at time $t$, $T_{t \to t-1}$ is the relative camera transformation, and $p_{t}$ denotes pixel coordinates in homogeneous form.

Since the rigid flow is computed between consecutive frames under a static-scene assumption, it cannot accurately model the motion of dynamic objects. In contrast, the optical flow reflects the true pixel-wise motion, including both static and non-static elements. Therefore, we detect dynamic regions by identifying pixels where the two flows differ significantly. Specifically, we define a binary motion mask $M_{\text{flow}}$ by thresholding the magnitude of the difference between the rigid and optical flows:
\begin{equation}
  M_{\text{flow}} = \lvert F_{t \to t-1} - \hat{F}_{t \to t-1} \rvert > \tau,
  \label{eq:flow_mask}
\end{equation}
where $\tau$ is set to the mean of the flow differences across the image. This mask highlights pixels that violate the rigid-motion assumption and are therefore likely to belong to dynamic objects.

Using the generated motion mask $M_{\text{flow}}$, we introduce an
isolated triplet loss.
In each $k\times k$ image patch, the central pixel is designated as the
anchor.  Neighboring pixels whose motion labels match that of the anchor form the positive set
$\mathcal{P}_i^{+}$, whereas pixels with different motion labels constitute the
negative set $\mathcal{P}_i^{-}$.  We compute the average positive and negative
distances as follows:
%수식 blue 처리가 안되어서 주석을 달아놓습니다. Triplet loss에 대한 수식입니다.
\begin{equation}
  D_i^{+} =
  \frac{1}{|\mathcal{P}_i^{+}|}
  \sum_{j \in \mathcal{P}_i^{+}}
  \bigl\| \hat f_i - \hat f_j \bigr\|_2^{2}.
  \label{eq:dplus}
\end{equation}

\begin{equation}
  D_i^{-} =
  \frac{1}{|\mathcal{P}_i^{-}|}
  \sum_{j \in \mathcal{P}_i^{-}}
  \bigl\| \hat f_i - \hat f_j \bigr\|_2^{2}.
  \label{eq:dminus}
\end{equation}

\noindent
Here, \(D_i^{+}\) denotes the anchor–positive distance and \(D_i^{-}\) the
anchor–negative distance, both defined as the mean squared Euclidean
distance between $\ell_2$-normalised depth features~\cite{chen2023self}.

% ---------- (2)  Isolated Triplet Loss ---------------------------
\begin{equation}
  L_{\text{tri}}
  = \frac{1}{|\Gamma|}
  \sum_{i\in\Gamma}
  \Bigl(
  D_i^{+} \;+\;
  \bigl[m_{0}-D_i^{-}\bigr]_{+}
  \Bigr),
  \label{eq:isolated_triplet_loss}
\end{equation}

\noindent
with \(f_u\) denoting the depth feature at pixel \(u\) and
\(\hat f_u = f_u/\lVert f_u\rVert_2\) its $\ell_2$-normalised feature,
\(\mathcal{P}_i^{+}\) is the set of pixels in the \(k\times k\) window centred at anchor \(i\) whose motion label is consistent with that of the anchor, whereas \(\mathcal{P}_i^{-}\) contains the pixels in the same window whose motion label is inconsistent,
\(\Gamma = \bigl\{\,i \;\bigm|\; \bigl(|\mathcal{P}_i^{+}| > k\bigr) \land \bigl(|\mathcal{P}_i^{-}| > k\bigr) \bigr\}\) denotes the boundary-anchor set,
\([\cdot]_+\) is the hinge operator, and the isolated margin is fixed to \(m_{0}=0.65\).

\subsection{Uncertainty Estimation}
Texture-less regions, such as the sky or plain walls, lack distinctive visual features or gradients, making it difficult to identify reliable correspondences across views. Consequently, when comparing the warped image $I_{s \to t}$, generated using incorrect depth or pose in such regions, with the target image $I_{t}$, the photometric error may be unintentionally computed as low. This can mislead the network into treating inaccurate depth predictions as correct, thereby reinforcing erroneous estimations. To mitigate this issue, we first estsimate the uncertainty.

Various studies ~\cite{kendall2017uncertainties, poggi2020uncertainty, li2023uncertainty} have proposed methods for incorporating uncertainty into depth estimation. Based on the approaches introduced in ~\cite{kendall2017uncertainties, li2023uncertainty}, we design the Student network to directly estimate the per-pixel variance $\sigma^2$ associated with its depth predictions.
% The estimated variance is then used to define the uncertainty, formulated as follows:
The estimated variance is then used to define the uncertainty associated with the predicted depth, formulated as follows:

\begin{equation}
  L_{\text{u}} = \frac{1}{N} \sum_{i=1}^{N} \left( \frac{L_{\text{ph, i}}^2}{\sigma_i^2} + \log(\sigma_i^2) \right)
  \label{eq:uncertainty_loss},
\end{equation}
where $N$ denotes the number of pixels. This formulation allows high-uncertainty regions to contribute less to the photometric error, while low-uncertainty regions have a greater impact. Using this property, an uncertainty mask is generated and leveraged to refine the depth estimation (Section~\ref{sec:Architecture} for more details).

Finally, while maintaining the baseline photometric loss, we incorporate the estimated uncertainty to formulate the final loss function as:
\begin{equation}
  L_{\text{total}} = \frac{1}{S} \sum_{i=0}^{S-1} (L_{\text{self}, \, i} + \lambda_{\text{u}} \cdot L_{\text{u}, \, i}+\lambda_{\text{tri}} \cdot L_{\text{tri}, \, i}).
  \label{eq:total_loss}
\end{equation}
Here, $S = 4$ denotes the number of multi-scale depth maps, where depth predictions are made at multiple resolutions to improve training stability. The uncertainty loss weight $\lambda_{\text{u}}$ is set to 1, and the triplet loss weight $\lambda_{\text{tri}}$ is set to 0.1.

\subsection{Model Architecture}
\label{sec:Architecture}

\subsubsection{Multi-View Mamba Encoder}
The overall structure of the proposed UM-Depth network is illustrated in Fig.~\ref{fig:um_model}. This network comprises a teacher network, which includes a convolution-based single-frame depth network and an optical flow network, and a student network, which consists of a multi-frame depth network.

In the student network, we adopt the approach from ~\cite{watson2021temporal} to construct a cost volume and integrate Mamba technique ~\cite{gu2023mamba} into the encoder, including the feature extractor. This enables efficient modeling of both local and global information while reducing computational complexity. Specifically, we employ GroupMamba ~\cite{shaker2024groupmamba}, which scans input images from multiple directions, effectively capturing a broader range of spatial information and efficiently integrating local and global features.

By leveraging the architectural distinction between the teacher and student networks, where the teacher provides stable, motion-aware supervision and the student exploits temporal context through multi-frame encoding, our method facilitates complementary learning that improves depth prediction accuracy.

\subsubsection{Prompt-HRNet}
\label{sec:prompt_hrnet}

Recent advancements in deep learning for image-based tasks have explored various techniques that utilize learnable parameters as prompts ~\cite{jia2022visual, wang2024tsp}. These prompts are trained alongside the model and contribute to performance improvements in various applications, such as denoising input data ~\cite{potlapalli2023promptir} and multimodal processing ~\cite{zhu2023visual}.

In this study, we adopt a prompt-based approach for depth estimation by extending the widely used HRNet architecture~\cite{sun2019deep}. This prompt-based component, which we refer to as the prompt branch of our PBU-HRNet decoder, is initialized with random parameters and contains no prior information. During training, however, it gradually learns to encode image-specific characteristics, thereby enabling richer depth feature extraction.

Without modifying the structural design of HRNet or introducing additional modules, we propose that simply concatenating the prompt with the intermediate feature maps along the channel dimension can enhance performance. Specifically, the intermediate feature $F$ generated by HRNet is concatenated with the prompt $P$ along the channel dimension to form $\tilde{F} = \text{Concat}(F, P)$, which is then forwarded to the subsequent layers to improve depth estimation performance. The overall structure of Prompt-HRNet is illustrated in Fig.~\ref{fig:um_model}.

\subsubsection{Bins-HRNet}
To improve the accuracy of depth estimation, we reformulate the conventional regression task as a classification problem by discretizing the depth range into predefined intervals using a binning strategy ~\cite{bhat2021adabins}. While previous methods ~\cite{bhat2021adabins, li2024binsformer} often introduce additional modules, our approach applies the binning technique by directly passing the output of the HRNet decoder through an multi-layer perceptron layer followed by a softmax to estimate the per-pixel probability map $p_{\text{pre}}$.

This probability map is then linearly combined with the bin centers $\{ c(b_i) \}$ to obtain the initial depth map $d_{\text{pre}}$. First, given a vector of bin widths $b$, the bin center $c(b_i)$ is defined as follows \cite{bhat2021adabins}:
\begin{equation}
  c(b_i) = d_{\min} + (d_{\max} - d_{\min}) \left( 2b_i + \sum_{j=1}^{i-1} b_j \right),
  \label{eq:bin_center}
\end{equation}
where $d_{\min}$ and $d_{\max}$ denote the minimum and maximum depth values, respectively.

The initial depth value $d_{\text{pre}}$ for each pixel is then computed as the weighted sum of bin centers using the estimated probability $p_{\text{pre}, i}$:
\begin{equation}
  D_{\text{pre}} = \sum_{i=1}^{N} c(b_i) \cdot p_{\text{pre}, i}
  \label{eq:depth_pre}.
\end{equation}

However, in low-texture regions such as the sky, the model may produce large errors. To address this, we utilize the uncertainty $\sigma^2$ estimated from the fusion module to generate an uncertainty mask $M_{\text{uncertainty}}$. Pixels with uncertainty above a threshold $\epsilon$ (defined as the top 20\% of $\sigma^2$) are identified as:
\begin{equation}
  M_{\text{u}} = \lvert\sigma^2\rvert > \epsilon,
  \label{eq:uncertainty_mask}
\end{equation}
where $\epsilon$ is the 80th percentile threshold of the estimated uncertainty values.

In high-uncertainty regions, the probability map $p_{\text{pre}}$ is refined to obtain a new distribution $p_{\text{post}}$, from which the refined depth map $d_{\text{post}}$ is computed using the same bin centers:
\begin{equation}
  D_{\text{post}} = \sum_{i=1}^{N} c(b_i) \cdot p_{\text{post}, i}
  \label{eq:depth_post}.
\end{equation}

Finally, the overall depth map $D_{t}$ is obtained by selectively using either the initial or refined depth values, depending on the uncertainty:

\begin{equation}
  D_{t}(x, y) =
  \begin{cases}
    D_{\text{post}}(x, y), & \text{if } M_{\text{u}}(x, y) = 1 \\
    D_{\text{pre}}(x, y),  & \text{otherwise}
  \end{cases}
  \label{eq:final_depth}
\end{equation}

Together with the prompt-based branch (Section~\ref{sec:prompt_hrnet}),
this bins branch forms our \emph{Prompt- and Bins-Uncertainty HRNet (PBU-HRNet) decoder},
which leverages both learned prompts and adaptive bins to refine depth predictions.

\begingroup          %  ⬇ tighten just for these two floats
\setlength{\textfloatsep}{6pt plus 1pt minus 1pt} % above 1st / below 2nd
\setlength{\floatsep}{4pt plus 1pt minus 1pt}     % between the two tables

% 선 두께·색, 간격 공통 매크로
\newcommand{\fixarraylines}{%
  \setlength{\arrayrulewidth}{1.0pt}% 모든 표 선 두께 (더욱 두껍게)
  \arrayrulecolor{black}%
  \doublerulesep=0pt                % double-rule 여백 제거
}
\newcommand{\tighttablesetup}{%
  \scriptsize                      % ≈ 7 pt: 폭 확보
  \setlength{\tabcolsep}{4pt}      % 열 간격
  \renewcommand{\arraystretch}{1.2}% 행 높이 약간 늘림
}
% ─────────────────────────────────────────────────────────────

% ── preamble 또는 표 위쪽에 추가 ─────────────────────────────
% “낮을수록 좋다”  ↓   /   “높을수록 좋다”  ↑
\newcommand{\betterdown}{\kern0.15em\scriptsize$\downarrow$}
\newcommand{\betterup}{\kern0.15em\scriptsize$\uparrow$}
% ────────────────────────────────────────────────────────────

% ─────────────── TABLE (adjustbox 없는 버전) ────────────────
\begin{table*}[t!]
  \centering
  \tighttablesetup    % ← 글꼴·간격 설정
  \fixarraylines      % ← 선 두께·색 고정

% \begin{table*}[t!]
%   \centering
  % \caption{result}
  \caption{Quantitative results on the KITTI dataset comparing self-supervised monocular depth estimation methods at two resolutions, where our method achieves the best overall performance.}

  \label{tab:kitti_results_main}

  \setlength{\tabcolsep}{4pt}
  \renewcommand{\arraystretch}{1.2}

  % \begin{adjustbox}{max width=\textwidth}
    \setlength{\arrayrulewidth}{1.0pt}   % 확실히 고정 (더욱 두껍게)
    \begin{tabular}{|l|c|c|c|c|c|c|c|c|c|c|c|}
      \hline
      \textbf{Method}                          & \textbf{M}               & \textbf{S}             & \textbf{F} & \textbf{W$\!\times\!$H} &
      \cellcolor{red!20}Abs Rel\betterdown &
      \cellcolor{red!20}Sq Rel\betterdown &
      \cellcolor{red!20}RMSE\betterdown &
      \cellcolor{red!20}RMSE log\betterdown &
      \cellcolor{blue!20}$\delta\!<\!1.25$\betterup &
      \cellcolor{blue!20}$\delta\!<\!1.25^{2}$\betterup &
      \cellcolor{blue!20}$\delta\!<\!1.25^{3}$\betterup                                                                                                                                                                                                            \\
      \hline
      Ranjan~\cite{ranjan2019competitive}                                   &                          &                        & $\bullet$  & 832$\times$256          & 0.148          & 1.149          & 5.464          & 0.226          & 0.815          & 0.935          & 0.973          \\
      EPC++~\cite{luo2019every}                                    &                          &                        & $\bullet$  & 832$\times$256          & 0.141          & 1.029          & 5.350          & 0.216          & 0.815          & 0.942          & 0.976          \\
      SC-Depth~\cite{bian2021unsupervised}                            &                &                        &  & 832$\times$256          & 0.114          & 0.813          & 4.706          & 0.191          & 0.873          & 0.960          & 0.982          \\
      Monodepth2~\cite{godard2019digging}                               &                          &                        &            & 640$\times$192          & 0.115          & 0.903          & 4.863          & 0.193          & 0.877          & 0.959          & 0.981          \\
      Packnet-SFM~\cite{guizilini20203d}                              &                          &                        &            & 640$\times$192          & 0.111          & 0.785          & 4.601          & 0.189          & 0.878          & 0.960          & 0.982          \\
      Patil~\cite{patil2020don}                                    & $\bullet$                &                        &            & 640$\times$192          & 0.111          & 0.821          & 4.650          & 0.187          & 0.883          & 0.961          & 0.982          \\
      HR-Depth~\cite{lyu2021hr}                                 &                          &                        &            & 640$\times$192          & 0.107          & 0.785          & 4.612          & 0.185          & 0.887          & 0.962          & 0.982          \\
      FeatDepth~\cite{shu2020feature}                                &                 &                        &   & 640$\times$192          & 0.104          & 0.729          & 4.481          & 0.179          & 0.893          & 0.965          & 0.984          \\
      DIFFNet~\cite{zhou2021self}                                  &                          &                        &            & 640$\times$192          & 0.102          & 0.764          & 4.483          & 0.180          & 0.896          & 0.965          & 0.983          \\
      Guizilini~\cite{guizilini2020semantically}                                &                          & $\bullet$                     &            & 640$\times$192          & 0.102          & 0.698          & 4.381          & 0.178          & 0.896          & 0.964          & 0.984          \\
      FSRE-Depth~\cite{jung2021fine}                               & $\bullet$                & $\bullet$              &            & 640$\times$192          & 0.102          & 0.675          & 4.393          & 0.178          & 0.893          & 0.966          & 0.984          \\
      MonoViT~\cite{zhao2022monovit}                                  &                          &                        &            & 640$\times$192          & 0.099          & 0.708          & 4.372          & 0.175          & 0.900          & 0.967          & 0.984          \\
      ManyDepth~\cite{watson2021temporal}                                & $\bullet$                &                        &            & 640$\times$192          & 0.098          & 0.770          & 4.459          & 0.176          & 0.900          & 0.965          & 0.983          \\
      DCPI-Depth~\cite{zhang2024dcpi}                               & $\bullet$                &                        & $\bullet$  & 640$\times$192          & 0.095          & \textbf{0.662}          & 4.274          & 0.170          & 0.902          & 0.967          & 0.985          \\
      TriDepth~\cite{chen2023self}                                 & $\bullet$                & $\bullet$              &            & 640$\times$192          & 0.093          & 0.665          & 4.272          & 0.172          & 0.907          & 0.967          & 0.984          \\
      \hline
      \cellcolor{black!8}\textbf{Ours}                            & \cellcolor{black!8}$\bullet$                & \cellcolor{black!8}                       & \cellcolor{black!8}$\bullet$  & \cellcolor{black!8}640$\times$192          & \cellcolor{black!8}\textbf{0.089} & \cellcolor{black!8}0.667 & \cellcolor{black!8}\textbf{4.147} & \cellcolor{black!8}\textbf{0.166} & \cellcolor{black!8}\textbf{0.915} & \cellcolor{black!8}\textbf{0.969} & \cellcolor{black!8}\textbf{0.985} \\
      \hline
      Packnet-SFM~\cite{guizilini20203d}                              &                          &                        &            & 1280$\times$384         & 0.107          & 0.802          & 4.538          & 0.186          & 0.889          & 0.962          & 0.981          \\
      Guizilini~\cite{guizilini2020semantically}                              &                          & $\bullet$                        &            & 1280$\times$384         & 0.100          & 0.761          & 4.270          & 0.175          & 0.902          & 0.965          & 0.982          \\
      Monodepth2~\cite{godard2019digging}                               &                          &                        &            & 1024$\times$320         & 0.115          & 0.882          & 4.701          & 0.190          & 0.879          & 0.961          & 0.982          \\
      DevNet~\cite{zhou2022devnet}                                   &                          &                        &            & 1024$\times$320         & 0.103          & 0.713          & 4.459          & 0.177          & 0.890          & 0.965          & 0.982          \\
      HR-Depth~\cite{lyu2021hr}                                 &                          &                        &            & 1024$\times$320         & 0.101          & 0.716          & 4.395          & 0.179          & 0.899          & 0.966          & 0.983          \\
      DIFFNet~\cite{zhou2021self}                                  &                          &                        &            & 1024$\times$320         & 0.097          & 0.722          & 4.345          & 0.174          & 0.907          & 0.967          & 0.984          \\
      MonoViT~\cite{zhao2022monovit}                                  &                          &                        &            & 1024$\times$320         & 0.094          & 0.682          & 4.200          & 0.170          & 0.912          & 0.969          & 0.984          \\
      ManyDepth~\cite{watson2021temporal}                                &$\bullet$                          &                        &            & 1024$\times$320         & 0.091          & 0.694          & 4.245          & 0.171          & 0.911          & 0.968          & 0.983          \\
      DCPI-Depth~\cite{zhang2024dcpi}                               &$\bullet$                          &                        &$\bullet$            & 1024$\times$320         & 0.090          & 0.655          & 4.113          & 0.167          & 0.914          & 0.969          & 0.985          \\
      \hline

      \cellcolor{black!8}\textbf{Ours}                            & \cellcolor{black!8}$\bullet$                & \cellcolor{black!8}                       & \cellcolor{black!8}$\bullet$  & \cellcolor{black!8}1024$\times$320         & \cellcolor{black!8}\textbf{0.089} & \cellcolor{black!8}\textbf{0.646} & \cellcolor{black!8}\textbf{4.088} & \cellcolor{black!8}\textbf{0.167} & \cellcolor{black!8}\textbf{0.913} & \cellcolor{black!8}\textbf{0.970} & \cellcolor{black!8}\textbf{0.985} \\
      \hline
    \end{tabular}
  % \end{adjustbox}

  \vspace{4pt}
  \begin{minipage}{\textwidth}
    \footnotesize
    \textbf{M} denotes multi–frame training, \textbf{S} indicates the use of semantic-segmentation guidance, and \textbf{F} refers to auxiliary optical-flow supervision.
    Results are reported in two resolution groups—$640\times192$ and $1024\times320$.
    The proposed method (\textbf{Ours}) is highlighted with a gray background for ease of comparison.
    Boldface indicates the best result within each resolution group.
  \end{minipage}
\end{table*}

\vspace{-6pt}           % (선택) 표 사이 여백 조정

\begin{table*}[!t]
  \centering
  \tighttablesetup   % 글꼴·간격
  \fixarraylines     % 선 두께·색
  \caption{Quantitative results on the Cityscapes dataset comparing self-supervised monocular depth estimation methods, where our method achieves the best overall performance.}
  \label{tab:city_results_main}

  \setlength{\tabcolsep}{4pt}
  \renewcommand{\arraystretch}{1.2}
  \setlength{\arrayrulewidth}{1.0pt}   % 확실히 고정 (더욱 두껍게)
  \begin{tabular}{|l|c|c|c|c|c|c|c|c|c|c|c|}
    \hline
    \textbf{Method} & \textbf{M} & \textbf{S} & \textbf{F} & \textbf{W$\!\times\!$H} &
    \cellcolor{red!20}Abs Rel\betterdown &
    \cellcolor{red!20}Sq Rel\betterdown &
    \cellcolor{red!20}RMSE\betterdown &
    \cellcolor{red!20}RMSE log\betterdown &
    \cellcolor{blue!20}$\delta\!<\!1.25$\betterup &
    \cellcolor{blue!20}$\delta\!<\!1.25^{2}$\betterup &
    \cellcolor{blue!20}$\delta\!<\!1.25^{3}$\betterup \\
    \hline
    % 640x192, 832x256, 1280x384 해상도 결과
    InstaDM~\cite{lee2021learning}                                  &                          & $\bullet$              &            & 832$\times$256               & 0.111          & 1.158          & 6.437          & 0.182          & 0.868          & 0.961          & 0.983          \\
    Pilzer~\cite{pilzer2018unsupervised}                                   &                          & $\bullet$              &            & 512$\times$256               & 0.240          & 4.264          & 8.049          & 0.334          & 0.710          & 0.871          & 0.937          \\
    Monodepth2~\cite{godard2019digging}                               &                          &                        &            & 416$\times$128               & 0.129          & 1.569          & 6.876          & 0.187          & 0.849          & 0.957          & 0.983          \\
    Struct2Depth~\cite{casser2019depth}                             & $\bullet$                &                        &            & 416$\times$128               & 0.151          & 2.492          & 7.024          & 0.202          & 0.826          & 0.937          & 0.972          \\
    Videos in the Wild~\cite{gordon2019depth}                       &                          &                        &            & 416$\times$128               & 0.127          & 1.330          & 6.960          & 0.195          & 0.830          & 0.947          & 0.981          \\
    Li~\cite{li2023learning}                                       &                          &                        &            & 416$\times$128               & 0.119          & 1.290          & 6.980          & 0.190          & 0.846          & 0.952          & 0.982          \\
    ManyDepth~\cite{watson2021temporal}                                & $\bullet$                &                        &            & 416$\times$128               & \textbf{0.114}          & 1.193          & 6.223          & 0.170          & 0.875          & 0.967          & 0.989          \\
    \hline

    \cellcolor{gray!20}Ours                                     & \cellcolor{gray!20}$\bullet$                & \cellcolor{gray!20}                       & \cellcolor{gray!20}$\bullet$  & \cellcolor{gray!20}416$\times$128               & \cellcolor{gray!20}0.116 & \cellcolor{gray!20}\textbf{1.007} & \cellcolor{gray!20}\textbf{5.872} & \cellcolor{gray!20}\textbf{0.163} & \cellcolor{gray!20}\textbf{0.877} & \cellcolor{gray!20}\textbf{0.974} & \cellcolor{gray!20}\textbf{0.992} \\
    \hline
  \end{tabular}

  \vspace{4pt}
  \begin{minipage}{\textwidth}
    \footnotesize
    \textbf{M} denotes multi–frame training, \textbf{S} indicates the use of semantic-segmentation guidance, and \textbf{F} refers to auxiliary optical-flow supervision.
    All results are evaluated on the Cityscapes dataset at a resolution of $416 \times 128$.
    The proposed method (\textbf{Ours}) is highlighted with a gray background for clarity.
    Boldface indicates the best performance among the compared methods.

  \end{minipage}
\end{table*}
\endgroup           %  ⬆ restore the original glue afterwards

\section{Experimental Results}
\subsection{Datasets}

% \subsubsection{Datasets}
To evaluate the performance of the proposed model, we utilize the KITTI Raw ~\cite{geiger2012we} and Cityscapes ~\cite{cordts2016cityscapes} datasets. The KITTI Raw dataset, widely used in autonomous driving and visual odometry tasks, consists of outdoor driving scenes. Following the protocol in ~\cite{zhou2017unsupervised}, we use 39810 images for training and 4424 images for validation. During training, identical intrinsic camera parameters are applied to all input images. For depth estimation evaluation, we adopt the Eigen split ~\cite{eigen2014depth}, comparing predictions against ground truth on a test set of 697 images.

The Cityscapes dataset ~\cite{cordts2016cityscapes}, comprising urban driving scenes, is also used for evaluation. Consistent with ~\cite{zhou2017unsupervised}, we use 69731 images for training and 1525 for testing. Evaluation metrics follow the protocol introduced in ~\cite{eigen2014depth}.

For camera pose estimation, we employ the KITTI Odometry benchmark ~\cite{geiger2013vision}, which contains 22 stereo sequences. Of these, 11 sequences include ground truth trajectories for training. Following the experimental settings in ~\cite{zhou2017unsupervised, zhan2018unsupervised}, sequences 00–08 are used for training and sequences 09–10 for validation. We evaluate the results using the average translation error \(e_t\),
the average rotation error \(e_r\), and the absolute trajectory error (ATE).

\subsection{Implementation Details}

Our model is implemented using PyTorch, and all experiments are conducted on a single NVIDIA RTX A100 GPU. Following ManyDepth ~\cite{watson2021temporal}, we train the network using the Adam optimizer ~\cite{kingma2014adam}. The learning rate is initially set to $1 \times 10^{-4}$ and reduced by a factor of 10 during the final five epochs. The batch size is set to 12. We train the network for 20 epochs on the KITTI and KITTI Odometry datasets, and for 5 epochs on the Cityscapes dataset.

We follow the previous works ~\cite{godard2017unsupervised, eigen2014depth} and evaluate the performance using standard metrics including Absolute Relative Error (Abs Rel), Squared Relative Error (Sq Rel), Root Mean Squared Error (RMSE), RMSE log, and accuracy under threshold $\delta < 1.25$, $\delta < 1.25^2$, and $\delta < 1.25^3$.

Unlike ManyDepth, which couples an ImageNet-pretrained ResNet18 encoder with the up-projection decoder of Godard et al.~\cite{godard2019digging}, our student network combines an ImageNet-pretrained ~\cite{deng2009imagenet} GroupMamba-Tiny encoder~\cite{shaker2024groupmamba} with a customized HRNet-based decoder, PBU-HRNet, that augments HRNet with learnable prompts, adaptive binning, and an uncertainty-guided refinement of high-uncertainty pixels~\cite{sun2019deep}.

Input resolutions are set to 192 $\times$ 640 and 320 $\times$ 1024 for the KITTI benchmark, 192 $\times$ 640 for KITTI Odometry, and 128 $\times$ 416 for Cityscapes.

For visual odometry evaluation, following~\cite{zhan2018unsupervised}, we use the average translational error $e_t$, average rotational error $e_r$, and Absolute Trajectory Error (ATE) in meters.

\subsection{Depth Estimation on Benchmarks}

\subsubsection{KITTI Dataset}

Table~\ref{tab:kitti_results_main} presents the quantitative results on the KITTI ~\cite{geiger2012we} dataset at two input resolutions: $192\times640$ and $320\times1024$. The proposed UM-Depth outperforms existing self-supervised depth estimation methods. Notably, during training, UM-Depth leverages dynamic object features extracted by the optical flow model RPKNet ~\cite{morimitsu2024recurrent}, which contributes to improved depth estimation accuracy.

Unlike previous approaches that commonly use convolution-based encoder or Vision Transformer-based encoder, UM-Depth employs GroupMamba ~\cite{shaker2024groupmamba}, a state-space model-based encoder, to achieve superior performance.

Furthermore, UM-Depth surpasses recent state-of-the-art methods such as TriDepth ~\cite{chen2023self}, which incorporates semantic segmentation, and DCPI-Depth ~\cite{zhang2024dcpi}, which utilizes optical flow. Our method achieves an Abs Rel of 0.089 and RMSE log of 0.166 at a resolution of $192\times640$, demonstrating its effectiveness.

\begin{figure*}[t]
  \centering
  % PDF 가 있으면 품질-용량이 가장 좋습니다.
  %   \includegraphics[width=\textwidth]{figure_depth_comparison.pdf}
  %
  % PDF 대신 PNG 를 쓰고 싶다면 아래 줄을 주석 해제하세요.
  \includegraphics[width=\textwidth]{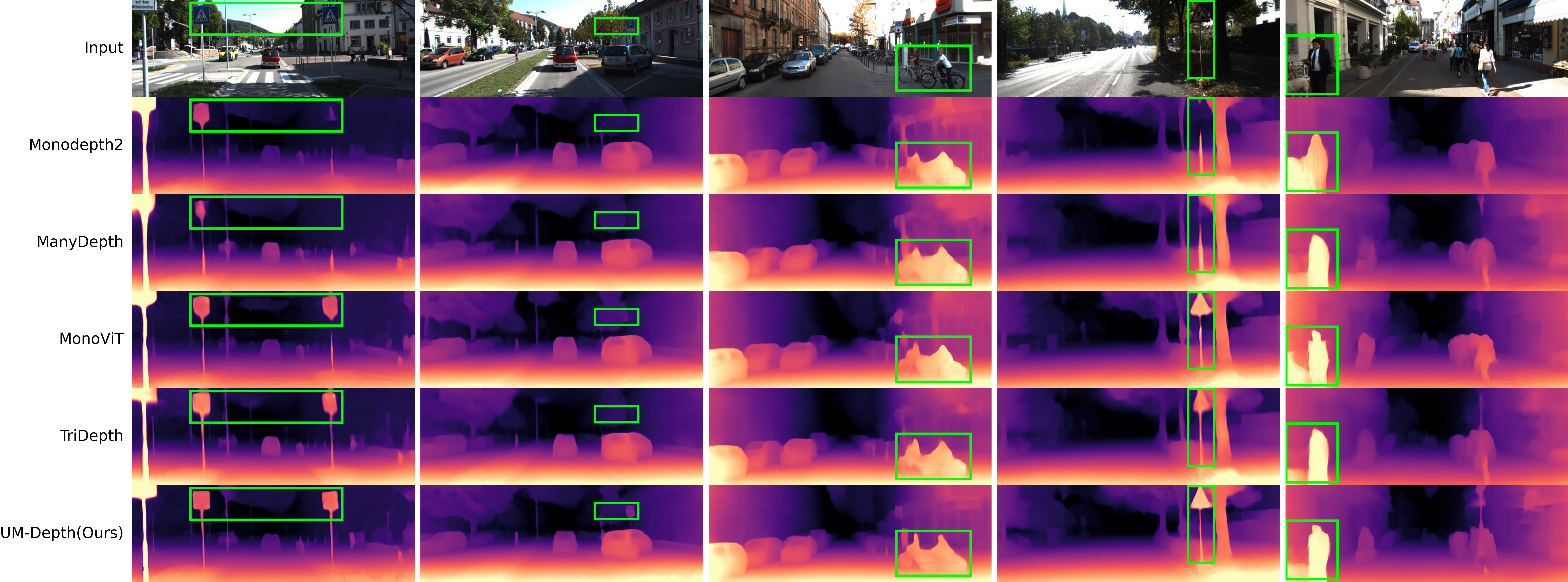}

  \caption{Qualitative results on the KITTI dataset. %
           Our approach produces higher–quality depth maps with finer object
           boundaries.}%
  \label{fig:kitti_qualitative}
\end{figure*}

Figure~\ref{fig:kitti_qualitative} illustrates qualitative results on the KITTI dataset. UM-Depth provides sharper and more detailed depth predictions, especially around object boundaries.

\subsubsection{Cityscapes Dataset}

Table ~\ref{tab:city_results_main} reports the performance of UM-Depth on the Cityscapes ~\cite{cordts2016cityscapes} dataset with an input resolution of $416\times128$. Compared with existing self-supervised methods such as ManyDepth ~\cite{watson2021temporal}, UM-Depth achieves competitive accuracy. In particular, it demonstrates strong generalization capability in complex urban scenes.
\setlength{\arrayrulewidth}{0.6pt}   % 0.6 pt 로 ↑
\arrayrulecolor{black}
\newcolumntype{C}{>{\centering\arraybackslash}X}
\doublerulesep=0pt
\begin{table}[t]
  \caption{Visual odometry results on the KITTI Odometry dataset; UM-Depth attains the lowest errors on Sequences 09 and 10.}
  \label{tab:kitti_vo_results}
  \centering
  \scriptsize \setlength{\tabcolsep}{2pt}\renewcommand{\arraystretch}{1.2}

  \begin{tabularx}{\columnwidth}{|l|CCC!{\vrule width 0.6pt}CCC|}
    \hline
    \multirow{2}{*}{\textbf{Method}} &
      \multicolumn{3}{c!{\vrule width 0.6pt}}{\cellcolor{green!20}\textbf{Seq.\,09}} &
      \multicolumn{3}{c|}{\cellcolor{green!20}\textbf{Seq.\,10}} \\ \cline{2-7}
    & $e_{t}$ (\%) & $e_{r}$ (\%) & ATE (m) &
      $e_{t}$ (\%) & $e_{r}$ (\%) & ATE (m) \\ \hline
    SfMLearner~\cite{zhou2017unsupervised}  & 19.15 & 6.82 & 77.79 & 40.40 & 17.69 & 67.34 \\ \hline
    GeoNet~\cite{yin2018geonet}      & 28.72 & 9.80 &158.45 & 23.90 & 9.00 & 43.04 \\ \hline
    DeepMatchVO~\cite{shen2019beyond} & 9.91 & 3.80 & 27.08 & 12.18 & 5.90 & 24.44 \\ \hline
    Monodepth2~\cite{godard2019digging}  & 36.70 &16.36 & 99.14 & 49.71 & 25.08 & 86.94 \\ \hline
    SC-Depth~\cite{bian2021unsupervised}    & 12.16 & 4.01 & 58.79 & 12.23 & 6.20 & 16.42 \\ \hline
    FeatDepth~\cite{shu2020feature}   & 8.75 & 2.11 & - & 10.67 & 4.91 & - \\ \hline
    \rowcolor{black!8}
    \textbf{UM-Depth (Ours)}
      & \textbf{4.90} & \textbf{1.75} & \textbf{12.47}
      & \textbf{7.49} & \textbf{2.33} & \textbf{11.93} \\
    \specialrule{\arrayrulewidth}{0pt}{0pt}
  \end{tabularx}
\end{table}

\begin{figure}[!t]
  \centering
  % 한 컬럼 너비에 맞추기
  \includegraphics[width=\linewidth]{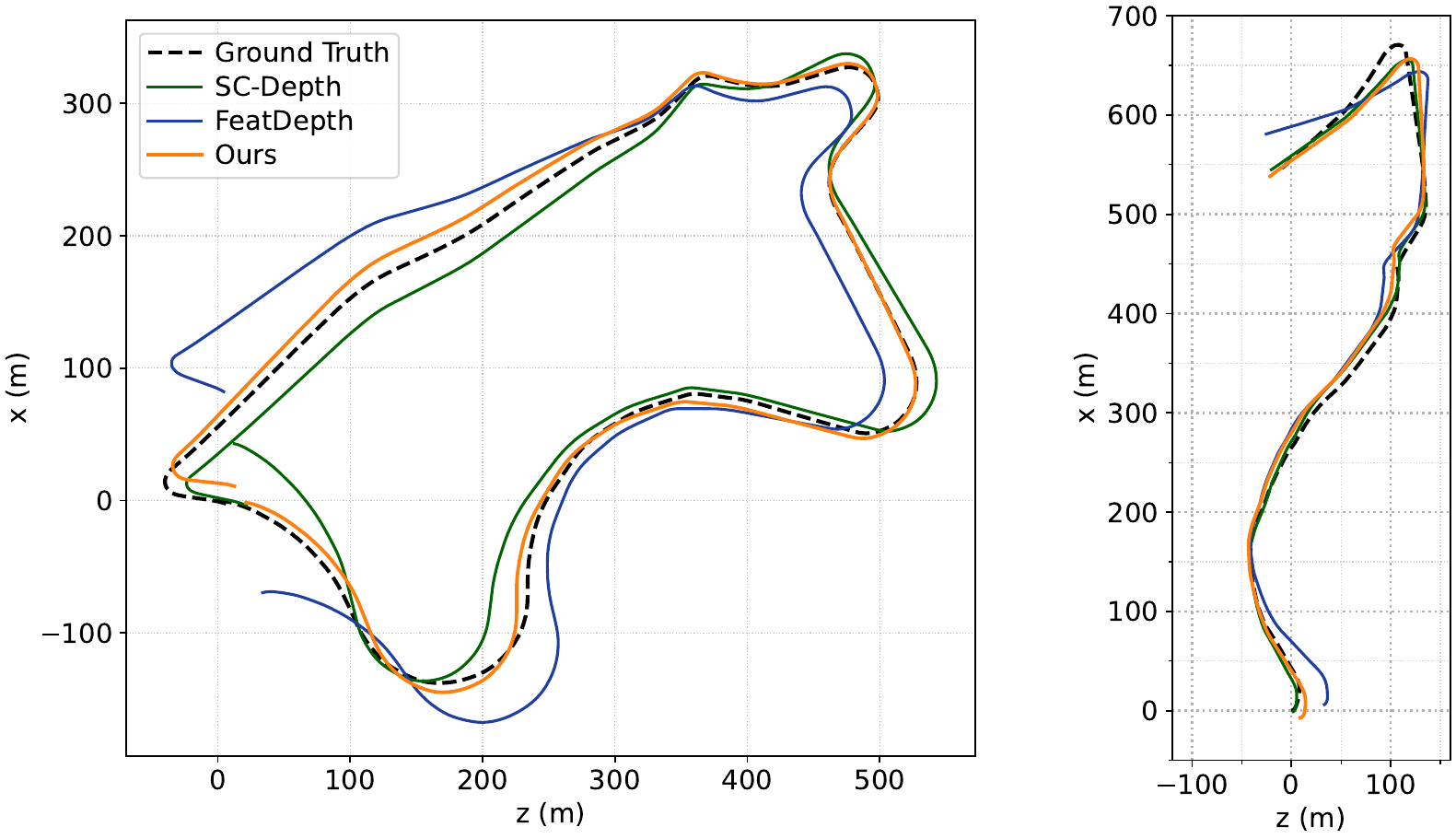}
  \vspace{-0.6cm}
  \caption{Qualitative comparison of predicted trajectories on Sequences 09 and 10 of the KITTI Odometry dataset~\cite{geiger2013vision}. All trajectories are aligned to the ground truth for fair visual comparison.}
  \label{fig:kitti_vo_seq0910}
  % \vspace{-1cm} % (선택) 위·아래 여백 미세 조정
\end{figure}

\subsection{Visual Odometry on Benchmarks}
\subsubsection{KITTI Odometry Dataset}

To evaluate the performance of our pose estimation network, we train and test on the KITTI Odometry dataset ~\cite{geiger2013vision}. Without introducing complex architectural changes, we adopt the widely used ResNet18 ~\cite{he2016deep} backbone while integrating GroupMamba ~\cite{shaker2024groupmamba}, a state-space model, as the encoder to efficiently process sequential image inputs and improve pose estimation accuracy.

Table~\ref{tab:kitti_vo_results} reports the root mean square error (RMSE) for translation and rotation compared to ground-truth trajectories. UM-Depth achieves superior performance over existing monocular visual odometry methods, demonstrating strong accuracy in both translational and rotational estimates. Compared to prior state-of-the-art frameworks, our approach achieves quantitatively better results across evaluated sequences.

Figure~\ref{fig:kitti_vo_seq0910} shows qualitative results for Sequences 09--10 from the KITTI Odometry dataset. Sequence 09 reflects trajectories in a complex road environment, while Sequence 10 captures motion in dense urban scenes. UM-Depth maintains stable performance across both scenarios and outperforms existing methods such as FeatDepth ~\cite{shu2020feature} and SC-Depth ~\cite{bian2021unsupervised}.
\subsection{Ablation Study}
To analyze the impact of each key component of the proposed UM-Depth on depth estimation performance, we conducted an ablation study using the KITTI dataset. All experiments were performed under the same training conditions with an input resolution of 192×640.

Table~\ref{tab:ablation_xyz} summarizes the quantitative results of the ablation study. When the optical flow-based Triplet Loss was removed, we observed a notable degradation in performance on the Abs Rel metric, indicating the effectiveness of this loss function in enhancing depth accuracy. Moreover, omitting uncertainty estimation and refinement, or removing the prompt module, resulted in significantly higher RMSE values. This demonstrates that both components contribute meaningfully to the stability and precision of depth prediction.

\begin{figure*}[!t]
  \centering
  % PDF 가 있으면 품질-용량이 가장 좋습니다.
  %   \includegraphics[width=\textwidth]{figure_depth_comparison.pdf}
  %
  % PDF 대신 PNG 를 쓰고 싶다면 아래 줄을 주석 해제하세요.
  \includegraphics[width=0.85\textwidth]{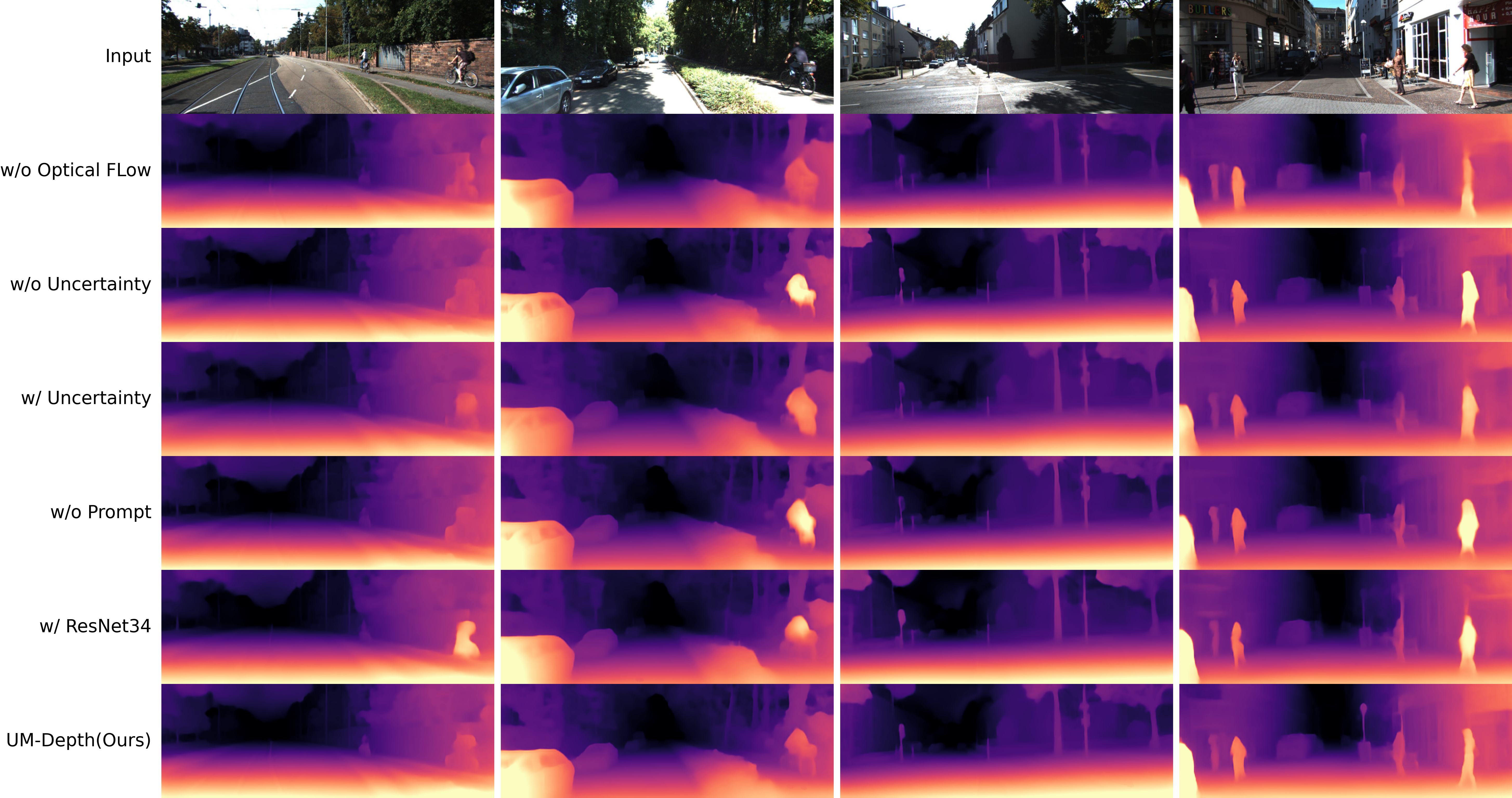}

  \caption{Qualitative depth results of our ablation study on the KITTI dataset.}
  \label{fig:kitti_Ablation}
  %\vspace{-2mm}  % ← 캡션과 본문 사이 공백을 조금 줄이고 싶을 때 사용
\end{figure*}

Finally, to assess the effectiveness of the Mamba-based encoder, we compared models using GroupMamba ~\cite{shaker2024groupmamba} and ResNet34 ~\cite{he2016deep}. The model with ResNet34 exhibited an overall performance drop, suggesting that the Mamba-based encoder enables more effective feature representation. Qualitative comparisons for each configuration are illustrated in Figure~\ref{fig:kitti_Ablation}.

% /////////////////////////////////////////////////////////////
%  Ablation table (replace the old one with this block)
% /////////////////////////////////////////////////////////////

\setlength{\arrayrulewidth}{0.45pt}
\arrayrulecolor{black}
% --------- 폭 유지하면서 간격까지 확보하는 매크로 ---------
\newcommand{\betterdown}{%
  \makebox[0.65em][l]{\kern0.12em
    {\scriptsize$\downarrow$}}%
}
\begin{table}[t]
  \caption{Ablation study on the KITTI dataset showing the impact of each component on depth estimation accuracy.}
  \label{tab:ablation_xyz}
  \centering
  \scriptsize
  \setlength{\tabcolsep}{5.5pt}
  \renewcommand{\arraystretch}{1.15}

  \begin{tabular}{|l|c|c|c|c|}
    \hline
    \textbf{Method} &
      \cellcolor{red!20}\textbf{Abs Rel\betterdown} &
      \cellcolor{red!20}\textbf{Sq Rel\betterdown}  &
      \cellcolor{red!20}\textbf{RMSE\betterdown}    &
      \cellcolor{red!20}\textbf{RMSElog\betterdown} \\ \hline
    w/o Optical Flow            & 0.094 & 0.645 & 4.161 & 0.170 \\ \hline
    w/o Uncertainty (no refine) & 0.091 & 0.702 & 4.227 & 0.171 \\ \hline
    w/ Uncertainty (no refine) & 0.091 & 0.681 & 4.216 & 0.167 \\ \hline
    w/o Prompt                  & 0.091 & 0.688 & 4.245 & 0.170 \\ \hline
    w/ ResNet34 Encoder         & 0.095 & 0.707 & 4.344 & 0.174 \\ \hline
    \rowcolor{black!8}
    \textbf{UM-Depth (Ours)}    & \textbf{0.089} & \textbf{0.667} & \textbf{4.147} & \textbf{0.166} \\ \hline
  \end{tabular}
\end{table}
\vspace{-2mm}  % ← 캡션과 본문 사이 공백을 조금 줄이고 싶을 때 사용

\FloatBarrier          % ★페이지를 넘기지 않고 여기서 즉시 floats 정리

\section{Conclusion}
In this paper, we proposed a novel pipeline for improving monocular depth estimation by leveraging an encoder based on Mamba. The proposed model effectively detects dynamic objects using optical flow information, thereby enhancing depth estimation accuracy around these regions. To further improve overall prediction quality, we introduced a prompt-based mechanism that guides the network, and we refined uncertain regions by quantitatively estimating uncertainty, resulting in more accurate depth maps. Extensive experiments conducted on the KITTI dataset demonstrated that our method achieves superior performance not only in depth estimation but also in odometry, outperforming existing state-of-the-art approaches. These results validate the effectiveness and practicality of the proposed framework.

% \vfill
% Make References
\printbibliography

@inproceedings{geiger2012we,
  title={Are we ready for autonomous driving? the kitti vision benchmark suite},
  author={Geiger, Andreas and Lenz, Philip and Urtasun, Raquel},
  booktitle = {{IEEE/CVF Conference on Computer Vision and Pattern Recognition (CVPR)}},
  pages={3354--3361},
  year={2012},
  % organization={IEEE}
}

@inproceedings{swan2006perceptual,
  title={A perceptual matching technique for depth judgments in optical, see-through augmented reality},
  author={Swan, J Edward and Livingston, Mark A and Smallman, Harvey S and Brown, Dennis and Baillot, Yohan and Gabbard, Joseph L and Hix, Deborah},
  booktitle = {{IEEE Virtual Reality Conference (VR 2006)}},
  pages={19--26},
  year={2006},
  % organization={IEEE}
}

@article{bonatto2021real,
  title={Real-time depth video-based rendering for 6-DoF HMD navigation and light field displays},
  author={Bonatto, Daniele and Fachada, Sarah and Rogge, S{\'e}gol{\`e}ne and Munteanu, Adrian and Lafruit, Gauthier},
  journal   = {{IEEE Access}},
  volume={9},
  pages={146868--146887},
  year={2021},
  publisher={IEEE}
}

@article{eigen2014depth,
  title={{Depth Map Prediction from a Single Image using a Multi-scale Deep Network}},
  author={Eigen, David and Puhrsch, Christian and Fergus, Rob},
  journal = {{Advances in Neural Information Processing Systems}},
  volume={27},
  year={2014}
}

@inproceedings{garg2016unsupervised,
  title={Unsupervised cnn for single view depth estimation: Geometry to the rescue},
  author={Garg, Ravi and Bg, Vijay Kumar and Carneiro, Gustavo and Reid, Ian},
  booktitle    = {{Computer Vision—ECCV 2016: 14th European Conference, Amsterdam, The Netherlands, October 11–14, 2016, Proceedings, Part VIII}},
  pages={740--756},
  year={2016},
  organization={Springer}
}

@inproceedings{zhou2017unsupervised,
  title={Unsupervised learning of depth and ego-motion from video},
  author={Zhou, Tinghui and Brown, Matthew and Snavely, Noah and Lowe, David G},
  booktitle = {{IEEE/CVF Conference on Computer Vision and Pattern Recognition (CVPR)}},
  pages={1851--1858},
  year={2017}
}

@inproceedings{godard2017unsupervised,
  title={Unsupervised monocular depth estimation with left-right consistency},
  author={Godard, Cl{\'e}ment and Mac Aodha, Oisin and Brostow, Gabriel J},
  booktitle = {{IEEE/CVF Conference on Computer Vision and Pattern Recognition (CVPR)}},
  pages={270--279},
  year={2017}
}

@inproceedings{zhan2018unsupervised,
  title={Unsupervised learning of monocular depth estimation and visual odometry with deep feature reconstruction},
  author={Zhan, Huangying and Garg, Ravi and Weerasekera, Chamara Saroj and Li, Kejie and Agarwal, Harsh and Reid, Ian},
  booktitle = {{IEEE/CVF Conference on Computer Vision and Pattern Recognition (CVPR)}},
  pages={340--349},
  year={2018}
}

@inproceedings{he2016deep,
  title={Deep residual learning for image recognition},
  author={He, Kaiming and Zhang, Xiangyu and Ren, Shaoqing and Sun, Jian},
  booktitle = {{IEEE/CVF Conference on Computer Vision and Pattern Recognition (CVPR)}},
  pages={770--778},
  year={2016}
}

@article{kendall2017uncertainties,
  title={What uncertainties do we need in bayesian deep learning for computer vision?},
  author={Kendall, Alex and Gal, Yarin},
  journal = {{Advances in Neural Information Processing Systems}},
  volume={30},
  year={2017}
}

@inproceedings{feng2022disentangling,
  title={Disentangling Object Motion and Occlusion for Unsupervised Multi-frame Monocular Depth},
  author={Feng, Ziyue and Yang, Liang and Jing, Longlong and Wang, Haiyan and Tian, YingLi and Li, Bing},
  booktitle = {{European Conference on Computer Vision (ECCV)}},
  pages={228--244},
  year={2022}
}

@article{wang2004image,
  title={Image quality assessment: from error visibility to structural similarity},
  author={Wang, Zhou and Bovik, Alan C and Sheikh, Hamid R and Simoncelli, Eero P},
  journal   = {{IEEE Transactions on Image Processing}},
  volume={13},
  number={4},
  pages={600--612},
  year={2004},
  publisher={IEEE}
}

@inproceedings{watson2021temporal,
  title={The temporal opportunist: Self-supervised multi-frame monocular depth},
  author={Watson, Jamie and Mac Aodha, Oisin and Prisacariu, Victor and Brostow, Gabriel and Firman, Michael},
  booktitle = {{IEEE/CVF Conference on Computer Vision and Pattern Recognition (CVPR)}},
  pages={1164--1174},
  year={2021}
}

@article{gu2023mamba,
  title={Mamba: Linear-time sequence modeling with selective state spaces},
  author={Gu, Albert and Dao, Tri},
  journal = {{arXiv preprint arXiv:2312.00752}},
  year={2023}
}

@article{shaker2024groupmamba,
  title={GroupMamba: Parameter-Efficient and Accurate Group Visual State Space Model},
  author={Shaker, Abdelrahman and Wasim, Syed Talal and Khan, Salman and Gall, Juergen and Khan, Fahad Shahbaz},
  journal = {{arXiv preprint arXiv:2407.13772}},
  year={2024}
}

@inproceedings{godard2019digging,
  title={Digging into self-supervised monocular depth estimation},
  author={Godard, Cl{\'e}ment and Mac Aodha, Oisin and Firman, Michael and Brostow, Gabriel J},
  booktitle = {{IEEE/CVF International Conference on Computer Vision (ICCV)}},
  pages={3828--3838},
  year={2019}
}

@article{zhang2024dcpi,
  title={DCPI-Depth: Explicitly infusing dense correspondence prior to unsupervised monocular depth estimation},
  author={Zhang, Mengtan and Feng, Yi and Chen, Qijun and Fan, Rui},
  journal = {{arXiv preprint arXiv:2405.16960}},
  year={2024}
}

@article{sun2024flowdepth,
  title={FlowDepth: Decoupling Optical Flow for Self-Supervised Monocular Depth Estimation},
  author={Sun, Yiyang and Xu, Zhiyuan and Wang, Xiaonian and Yao, Jing},
  journal = {{arXiv preprint arXiv:2403.19294}},
  year={2024}
}

@inproceedings{chen2023self,
  title={Self-supervised monocular depth estimation: Solving the edge-fattening problem},
  author={Chen, Xingyu and Zhang, Ruonan and Jiang, Ji and Wang, Yan and Li, Ge and Li, Thomas H},
  booktitle = {{IEEE/CVF Winter Conference on Applications of Computer Vision (WACV)}},
  pages={5776--5786},
  year={2023}
}

@inproceedings{jung2021fine,
  title={Fine-grained semantics-aware representation enhancement for self-supervised monocular depth estimation},
  author={Jung, Hyunyoung and Park, Eunhyeok and Yoo, Sungjoo},
  booktitle = {{IEEE/CVF International Conference on Computer Vision (ICCV)}},
  pages={12642--12652},
  year={2021}
}

@inproceedings{morimitsu2024recurrent,
  title={Recurrent partial kernel network for efficient optical flow estimation},
  author={Morimitsu, Henrique and Zhu, Xiaobin and Ji, Xiangyang and Yin, Xu-Cheng},
  booktitle = {{AAAI Conference on Artificial Intelligence (AAAI)}},
  volume={38},
  number={5},
  pages={4278--4286},
  year={2024}
}

@inproceedings{poggi2020uncertainty,
  title={On the uncertainty of self-supervised monocular depth estimation},
  author={Poggi, Matteo and Aleotti, Filippo and Tosi, Fabio and Mattoccia, Stefano},
  booktitle = {{IEEE/CVF Conference on Computer Vision and Pattern Recognition (CVPR)}},
  pages={3227--3237},
  year={2020}
}

@inproceedings{li2023uncertainty,
  title={Uncertainty Guided Self-Supervised Monocular Depth Estimation Based on Monte Carlo Method},
  author={Li, Ran and Liu, Zhong and Wu, Xingming and Liu, Jingmeng and Chen, Weihai},
  booktitle = {{2023 IEEE 18th Conference on Industrial Electronics and Applications (ICIEA)}},
  pages={90--95},
  year={2023},
  organization={IEEE}
}

@inproceedings{sun2019deep,
  title={Deep high-resolution representation learning for human pose estimation},
  author={Sun, Ke and Xiao, Bin and Liu, Dong and Wang, Jingdong},
  booktitle = {{IEEE/CVF Conference on Computer Vision and Pattern Recognition (CVPR)}},
  pages={5693--5703},
  year={2019}
}

@inproceedings{jia2022visual,
  title={Visual prompt tuning},
  author={Jia, Menglin and Tang, Luming and Chen, Bor-Chun and Cardie, Claire and Belongie, Serge and Hariharan, Bharath and Lim, Ser-Nam},
  booktitle = {{European Conference on Computer Vision (ECCV)}},
  pages={709--727},
  year={2022},
  organization={Springer}
}

@article{zhou2022learning,
  title={Learning to prompt for vision-language models},
  author={Zhou, Kaiyang and Yang, Jingkang and Loy, Chen Change and Liu, Ziwei},
  journal = {{International Journal of Computer Vision}},
  volume={130},
  number={9},
  pages={2337--2348},
  year={2022},
  publisher={Springer}
}

@inproceedings{radford2021learning,
  title={Learning transferable visual models from natural language supervision},
  author={Radford, Alec and Kim, Jong Wook and Hallacy, Chris and Ramesh, Aditya and Goh, Gabriel and Agarwal, Sandhini and Sastry, Girish and Askell, Amanda and Mishkin, Pamela and Clark, Jack and others},
  booktitle = {{International Conference on Machine Learning (ICML)}},
  pages={8748--8763},
  year={2021},
  organization={PmLR}
}

@article{potlapalli2023promptir,
  title={Promptir: Prompting for all-in-one image restoration},
  author={Potlapalli, Vaishnav and Zamir, Syed Waqas and Khan, Salman H and Shahbaz Khan, Fahad},
  journal = {{Advances in Neural Information Processing Systems}},
  volume={36},
  pages={71275--71293},
  year={2023}
}

@inproceedings{zhu2023visual,
  title={Visual prompt multi-modal tracking},
  author={Zhu, Jiawen and Lai, Simiao and Chen, Xin and Wang, Dong and Lu, Huchuan},
  booktitle = {{IEEE/CVF Conference on Computer Vision and Pattern Recognition (CVPR)}},
  pages={9516--9526},
  year={2023}
}

@inproceedings{wang2024tsp,
  title={Tsp-transformer: Task-specific prompts boosted transformer for holistic scene understanding},
  author={Wang, Shuo and Li, Jing and Zhao, Zibo and Lian, Dongze and Huang, Binbin and Wang, Xiaomei and Li, Zhengxin and Gao, Shenghua},
  booktitle = {{IEEE/CVF Winter Conference on Applications of Computer Vision (WACV)}},
  pages={925--934},
  year={2024}
}

@inproceedings{bhat2021adabins,
  title={Adabins: Depth estimation using adaptive bins},
  author={Bhat, Shariq Farooq and Alhashim, Ibraheem and Wonka, Peter},
  booktitle = {{IEEE/CVF Conference on Computer Vision and Pattern Recognition (CVPR)}},
  pages={4009--4018},
  year={2021}
}

@article{li2024binsformer,
  title={Binsformer: Revisiting adaptive bins for monocular depth estimation},
  author={Li, Zhenyu and Wang, Xuyang and Liu, Xianming and Jiang, Junjun},
  journal = {{IEEE Transactions on Image Processing}},
  year={2024},
  publisher={IEEE}
}

@article{liu2024vmamba,
  title={Vmamba: Visual state space model},
  author={Liu, Yue and Tian, Yunjie and Zhao, Yuzhong and Yu, Hongtian and Xie, Lingxi and Wang, Yaowei and Ye, Qixiang and Jiao, Jianbin and Liu, Yunfan},
  journal = {{Advances in Neural Information Processing Systems}},
  volume={37},
  pages={103031--103063},
  year={2024}
}

@article{hatamizadeh2024mambavision,
  title={Mambavision: A hybrid mamba-transformer vision backbone},
  author={Hatamizadeh, Ali and Kautz, Jan},
  journal = {{arXiv preprint arXiv:2407.08083}},
  year={2024}
}

@inproceedings{liu2024swin,
  title={Swin-umamba: Mamba-based unet with imagenet-based pretraining},
  author={Liu, Jiarun and Yang, Hao and Zhou, Hong-Yu and Xi, Yan and Yu, Lequan and Li, Cheng and Liang, Yong and Shi, Guangming and Yu, Yizhou and Zhang, Shaoting and others},
  booktitle = {{International Conference on Medical Image Computing and Computer-Assisted Intervention}},
  pages={615--625},
  year={2024},
  organization={Springer}
}

@inproceedings{cordts2016cityscapes,
  title={The cityscapes dataset for semantic urban scene understanding},
  author={Cordts, Marius and Omran, Mohamed and Ramos, Sebastian and Rehfeld, Timo and Enzweiler, Markus and Benenson, Rodrigo and Franke, Uwe and Roth, Stefan and Schiele, Bernt},
  booktitle = {{IEEE/CVF Conference on Computer Vision and Pattern Recognition (CVPR)}},
  pages={3213--3223},
  year={2016}
}

@article{kingma2014adam,
  title={Adam: A method for stochastic optimization},
  author={Kingma, Diederik P and Ba, Jimmy},
  journal = {{arXiv preprint arXiv:1412.6980}},
  year={2014}
}

@inproceedings{deng2009imagenet,
  title={Imagenet: A large-scale hierarchical image database},
  author={Deng, Jia and Dong, Wei and Socher, Richard and Li, Li-Jia and Li, Kai and Fei-Fei, Li},
  booktitle = {{2009 IEEE/CVF Conference on Computer Vision and Pattern Recognition (CVPR)}},
  pages={248--255},
  year={2009},
  organization={Ieee}
}

@article{geiger2013vision,
  title={Vision meets robotics: The kitti dataset},
  author={Geiger, Andreas and Lenz, Philip and Stiller, Christoph and Urtasun, Raquel},
  journal = {{The International Journal of Robotics Research}},
  volume={32},
  number={11},
  pages={1231--1237},
  year={2013},
  publisher={Sage Publications Sage UK: London, England}
}

@inproceedings{shu2020feature,
  title={Feature-metric loss for self-supervised learning of depth and egomotion},
  author={Shu, Chang and Yu, Kun and Duan, Zhixiang and Yang, Kuiyuan},
  booktitle = {{European Conference on Computer Vision (ECCV)}},
  pages={572--588},
  year={2020},
  organization={Springer}
}

@article{bian2021unsupervised,
  title={Unsupervised scale-consistent depth learning from video},
  author={Bian, Jia-Wang and Zhan, Huangying and Wang, Naiyan and Li, Zhichao and Zhang, Le and Shen, Chunhua and Cheng, Ming-Ming and Reid, Ian},
  journal = {{International Journal of Computer Vision}},
  volume={129},
  number={9},
  pages={2548--2564},
  year={2021},
  publisher={Springer}
}

@inproceedings{ranjan2019competitive,
  title={Competitive collaboration: Joint unsupervised learning of depth, camera motion, optical flow and motion segmentation},
  author={Ranjan, Anurag and Jampani, Varun and Balles, Lukas and Kim, Kihwan and Sun, Deqing and Wulff, Jonas and Black, Michael J},
  booktitle={{IEEE/CVF Conference on Computer Vision and Pattern Recognition (CVPR)}},
  pages={12240--12249},
  year={2019}
}

@article{luo2019every,
  title={Every pixel counts++: Joint learning of geometry and motion with 3D holistic understanding},
  author={Luo, Chenxu and Yang, Zhenheng and Wang, Peng and Wang, Yang and Xu, Wei and Nevatia, Ram and Yuille, Alan},
  journal={{IEEE Transactions on Pattern Analysis and Machine Intelligence}},
  volume={42},
  number={10},
  pages={2624--2641},
  year={2019},
  publisher={IEEE}
}

@inproceedings{guizilini20203d,
  title={3D packing for self-supervised monocular depth estimation},
  author={Guizilini, Vitor and Ambrus, Rares and Pillai, Sudeep and Raventos, Allan and Gaidon, Adrien},
  booktitle={{IEEE/CVF Conference on Computer Vision and Pattern Recognition (CVPR)}},
  pages={2485--2494},
  year={2020}
}

@article{patil2020don,
  title={Don't forget the past: Recurrent depth estimation from monocular video},
  author={Patil, Vaishakh and Van Gansbeke, Wouter and Dai, Dengxin and Van Gool, Luc},
  journal={{IEEE Robotics and Automation Letters}},
  volume={5},
  number={4},
  pages={6813--6820},
  year={2020},
  publisher={IEEE}
}

@inproceedings{lyu2021hr,
  title={HR-Depth: High resolution self-supervised monocular depth estimation},
  author={Lyu, Xiaoyang and Liu, Liang and Wang, Mengmeng and Kong, Xin and Liu, Lina and Liu, Yong and Chen, Xinxin and Yuan, Yi},
  booktitle={{AAAI Conference on Artificial Intelligence (AAAI)}},
  volume={35},
  number={3},
  pages={2294--2301},
  year={2021}
}

@article{zhou2021self,
  title={Self-supervised monocular depth estimation with internal feature fusion},
  author={Zhou, Hang and Greenwood, David and Taylor, Sarah},
  journal={{arXiv preprint arXiv:2110.09482}},
  year={2021}
}

@article{guizilini2020semantically,
  title={Semantically-guided representation learning for self-supervised monocular depth},
  author={Guizilini, Vitor and Hou, Rui and Li, Jie and Ambrus, Rares and Gaidon, Adrien},
  journal={{arXiv preprint arXiv:2002.12319}},
  year={2020}
}

@inproceedings{zhao2022monovit,
  title={MonoViT: Self-supervised monocular depth estimation with a vision transformer},
  author={Zhao, Chaoqiang and Zhang, Youmin and Poggi, Matteo and Tosi, Fabio and Guo, Xianda and Zhu, Zheng and Huang, Guan and Tang, Yang and Mattoccia, Stefano},
  booktitle={{International Conference on 3D Vision (3DV)}},
  pages={668--678},
  year={2022},
  organization={IEEE}
}

@inproceedings{zhou2022devnet,
  title={DevNet: Self-supervised monocular depth learning via density volume construction},
  author={Zhou, Kaichen and Hong, Lanqing and Chen, Changhao and Xu, Hang and Ye, Chaoqiang and Hu, Qingyong and Li, Zhenguo},
  booktitle={{European Conference on Computer Vision (ECCV)}},
  pages={125--142},
  year={2022},
  organization={Springer}
}

@inproceedings{lee2021learning,
  title={Learning monocular depth in dynamic scenes via instance-aware projection consistency},
  author={Lee, Seokju and Im, Sunghoon and Lin, Stephen and Kweon, In So},
  booktitle={{AAAI Conference on Artificial Intelligence (AAAI)}},
  volume={35},
  number={3},
  pages={1863--1872},
  year={2021}
}

@inproceedings{pilzer2018unsupervised,
  title={Unsupervised adversarial depth estimation using cycled generative networks},
  author={Pilzer, Andrea and Xu, Dan and Puscas, Mihai and Ricci, Elisa and Sebe, Nicu},
  booktitle={{International Conference on 3D Vision (3DV)}},
  pages={587--595},
  year={2018},
  organization={IEEE}
}

@inproceedings{casser2019depth,
  title={Depth prediction without the sensors: Leveraging structure for unsupervised learning from monocular videos},
  author={Casser, Vincent and Pirk, Soeren and Mahjourian, Reza and Angelova, Anelia},
  booktitle={{AAAI Conference on Artificial Intelligence (AAAI)}},
  volume={33},
  number={1},
  pages={8001--8008},
  year={2019}
}

@inproceedings{gordon2019depth,
  title={Depth from videos in the wild: Unsupervised monocular depth learning from unknown cameras},
  author={Gordon, Ariel and Li, Hanhan and Jonschkowski, Rico and Angelova, Anelia},
  booktitle={Proceedings of the IEEE/CVF international conference on computer vision},
  pages={8977--8986},
  year={2019}
}

@article{li2023learning,
  title={Learning depth via leveraging semantics: Self-supervised monocular depth estimation with both implicit and explicit semantic guidance},
  author={Li, Rui and Xue, Danna and Su, Shaolin and He, Xiantuo and Mao, Qing and Zhu, Yu and Sun, Jinqiu and Zhang, Yanning},
  journal={{Pattern Recognition}},
  volume={137},
  pages={109297},
  year={2023},
  publisher={Elsevier}
}

@inproceedings{yin2018geonet,
  title={GeoNet: Unsupervised learning of dense depth, optical flow and camera pose},
  author={Yin, Zhichao and Shi, Jianping},
  booktitle={{IEEE Conference on Computer Vision and Pattern Recognition (CVPR)}},
  pages={1983--1992},
  year={2018}
}

@inproceedings{shen2019beyond,
  title={Beyond photometric loss for self-supervised ego-motion estimation},
  author={Shen, Tianwei and Luo, Zixin and Zhou, Lei and Deng, Hanyu and Zhang, Runze and Fang, Tian and Quan, Long},
  booktitle={{International Conference on Robotics and Automation (ICRA)}},
  pages={6359--6365},
  year={2019},
  organization={IEEE}
}

% \IEEEtriggeratref{3} % equalize last page column height by dividing bib, at specific index number. It should be manually selected. IEEE_HowTo.
\end{document}